\documentclass{bmvc2k}

\usepackage{bbm}  
\usepackage{booktabs} 
\usepackage{multirow} 
\usepackage{xspace} 
\usepackage{enumitem}

\usepackage{graphicx} 
\usepackage{caption}
\usepackage{float}
\usepackage{wrapfig}

\usepackage{amsfonts}  
\usepackage{xpatch}  
\usepackage{xcolor,colortbl}
\usepackage{array, boldline, makecell} 
\usepackage{arydshln}  
\usepackage{soul} 
\usepackage{stackengine} 
\usepackage{algorithm}
\usepackage{algorithmic}

\usepackage{pifont}%
\newcommand{\cmark}{\ding{51}}
\newcommand{\xmark}{\ding{55}}

\newcommand{\thing}{\textit{thing}\xspace}
\newcommand{\stuff}{\textit{stuff}\xspace}

\newcommand{\unknown}{\textit{unknown}\xspace}
\newcommand{\known}{\textit{known}\xspace}

\newcommand{\unseen}{\textit{unseen}\xspace}

\newcommand{\val}{\textit{val}\xspace}

\title{Dual Decision Improves Open-Set Panoptic Segmentation}

\addauthor{Hai-Ming Xu}{hai-ming.xu@adelaide.edu.au}{1}
\addauthor{Hao Chen}{stanzju@gmail.com}{2}
\addauthor{Lingqiao Liu}{lingqiao.liu@adelaide.edu.au}{1}
\addauthor{Yufei Yin}{yinyufei@mail.ustc.edu.cn}{3}

\addinstitution{
 The University of Adelaide, \\
 Adelaide, Australia
}
\addinstitution{
 Zhejiang University\\
 Zhejiang, China
}
\addinstitution{
 University of Science and Technology of China\\
 Anhui, China
}

\runninghead{Hai-Ming Xu et al.}{OpenPS}

\begin{document}

\maketitle

\begin{abstract}
Open-set panoptic segmentation (OPS) problem is a new research direction aiming to perform segmentation for both \known classes and \unknown classes, i.e., the objects (``things'') that are never annotated in the training set. The main challenges of OPS are twofold: (1) the infinite possibility of the \unknown object appearances makes it difficult to model them from a limited number of training data. (2) at training time, we are only provided with the ``void'' category, which essentially mixes the ``unknown thing'' and ``background'' classes. We empirically find that directly using ``void'' category to supervise \known class or ``background'' classifiers without screening will lead to an unsatisfied OPS result. In this paper, we propose a divide-and-conquer scheme to develop a dual decision process for OPS. We show that by properly combining a \known class discriminator with an additional class-agnostic object prediction head, the OPS performance can be significantly improved. Specifically, we first propose to create a classifier with only \known categories and let the ``void'' class proposals achieve low prediction probability from those categories. Then we distinguish the ``unknown things'' from the background by using the additional object prediction head. To further boost performance, we introduce ``unknown things''  pseudo-labels generated from up-to-date models to enrich the training set. Our extensive experimental evaluation shows that our approach significantly improves \unknown class panoptic quality, with more than 30\% relative improvements than the existing best-performed method.
\end{abstract}

\begin{figure*}[t]\footnotesize
\centering
\setlength{\abovecaptionskip}{0.1cm}
\includegraphics[scale=0.35]{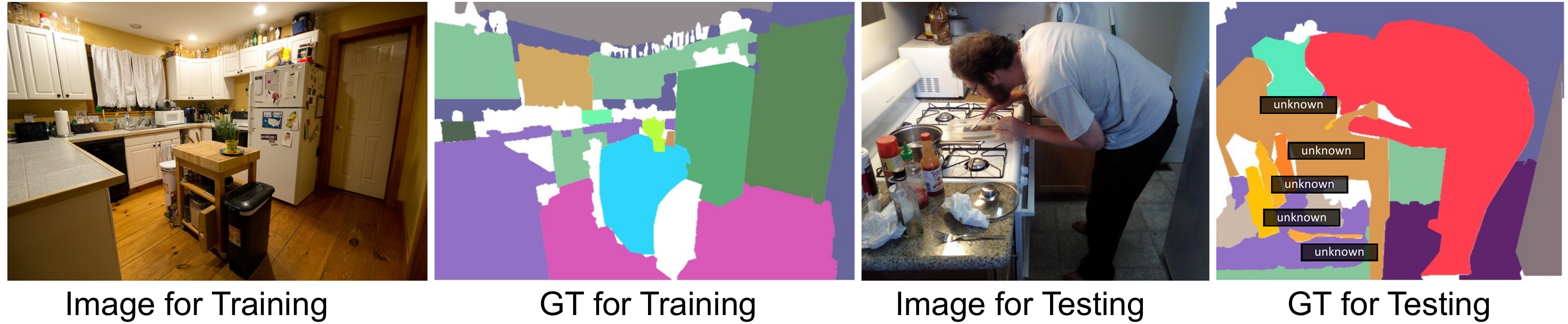}
\caption{Demonstration of OPS task setting. At training phase, an natural image and its ground truth segments (i.e., \known class \thing and \stuff) of interest are given for model learning. Some \thing objects that are not interesting or difficult to label are left as \emph{void}, i.e., blank area. While the model is required to be able to segment not only \known class objects, but also \unknown class objects (denoted by orange color in Testing GT) for testing images, e.g., bottles, fork and toothbrush.
}
\label{fig:ops_setting}
\end{figure*}
\section{Introduction}\label{sec:intro}
Panoptic segmentation (PS) is attracting growing attention from the vision community since it was proposed by Kirillov et al.~\cite{kirillov2019panoptic}. Such hot trend attributes to its ambitious goal for accommodating both semantic segmentation and instance segmentation in an unified framework and producing holistic scene parsing results~\cite{xiong2019upsnet,li2019attention,kirillov2019panoptic,cheng2020panoptic,wang2020axial,wang2021max,li2021fully,zhang2021k,cheng2021per,cheng2021masked,li2021panoptic}.
Most of the researches are built under a common closed-set assumption, i.e., the model only needs to segment the same class of objects appeared in the training set. However, such kind of systems will not be competent for the complex open-set scenario. For example, automatic driving can not identify abnormal objects will lead to catastrophic danger~\cite{abraham2016autonomous} and possible problems can even not be predictable in medical diagnosis~\cite{bakator2018deep}. Therefore, PS systems for dealing with open-set challenge are urgently demanded.

Open-set problem~\cite{scheirer2012toward,scheirer2014probability,geng2020recent,vaze2021open} has been well explored in classification tasks which refer to the scenario that when some new classes unseen in training appear in testing, the recognition model is required to not only accurately classify known classes given in training but also effectively deal with the unknown classes.

The recent research~\cite{hwang2021exemplar} extends PS to a realistic setting and firstly defines the \textit{open-set panoptic segmentation} (OPS) task. OPS takes categories given during training as \known classes and requires the model to produce segments for both \known and \unknown class objects (``things'') at testing phase, where the \unknown classes are never annotated or even appeared in the training set. As examples shown in Figure~\ref{fig:ops_setting}, many bottles in training image (above the closet) are hard to be labeled pixel by pixel and given as the \emph{void} area in ground truth segments. While for testing images, it is required to predict segments for different kinds of bottles and even the fork and toothbrush which are never appeared during training.

The OPS task is challenging because, on the one hand, the appearance of \unknown class objects are diverse and it would be hard for directly modeling \unknown classes from the given training images. On the other hand, although the ``void'' category is available at training phase, the training samples in ``void'' category are too noisy to provide effective supervisions since the ``unknown thing'' class and the ``background'' are confounded together. 

In order to tackle these two challenges, we choose to recognize the \unknown class objects in a dual decision process. Through coupling the \known class discriminator with an class-agnostic object prediction head, we can significantly improve the performance for the OPS task. Specifically, we build up a \known class classifier and suppress its predictions for ``void'' class proposals to compact the decision boundaries of \known classes and empower the \known class classifier the ability to reject non-known classes. Then, we further create a class-agnostic object prediction head to further distinguish ``unknown things'' from the background.
Moreover, we propose to use the pseudo-labeling method to further boost the generalization ability of the newly added object prediction head.
Extensive experimental results show that our approach has successfully achieved a new state-of-the-art performance on various OPS tasks.

\section{Related Work}\label{sec:related_work}

\smallskip
\noindent \textbf{Panoptic Segmentation} Pursuing a wholistic scene parsing, 
panoptic segmentation task (PS) is proposed to expect the generation of both semantic and instance segmentation simultaneously.
Given fully annotations to the training images, different kinds of modeling targets have been explored for the PS problem. Specifically, 
unified end-to-end networks~\cite{xiong2019upsnet,li2019attention,kirillov2019panoptic} are soon proposed after the initial release of baseline method with separate networks.
DeepLab series methods~\cite{cheng2020panoptic,wang2020axial,wang2021max} are deployed for fast inference speed.
More recently, universal image segmentation~\cite{li2021fully,zhang2021k,cheng2021per,cheng2021masked,li2021panoptic} is pursued.
While OPS shares a distinct target which demands the model to produce segments for \unknown classes that are never acknowledged during training.

\smallskip
\noindent \textbf{Open-Set Learning} Open-set problem has been well explored in the recognition/classification task~\cite{scheirer2012toward,scheirer2014probability,bendale2016towards,yoshihashi2019classification,oza2019c2ae,geng2020recent,vaze2021open}. The target of open-set recognition is to make the model successfully identify known classes and have the ability to identify unknown classes which are never exposed during training. OPS can be more challenging because \unknown classes are not provided intact, but needs to be detected by the model itself. Other related work includes open-world object detection~\cite{joseph2021towards,gupta2022ow} and open-world entity segmentation~\cite{qi2021open}. According to the problem definition, the former work contains a human labeling process after the \unknown class detection while OPS does not require. 
And the open-set recognition procedure proposed in OW-DETR~\cite{gupta2022ow} significantly differs from our approach, e.g., the generation and usage of pseudo unknown objects varies greatly and the unknown class decision process is also different.
While the latter one aims to segment visual entities without considering classes which is precisely the problem that OPS needs to solve urgently.

\begin{figure*}[t]\footnotesize
\centering
\setlength{\abovecaptionskip}{0.cm}
\includegraphics[scale=0.33]{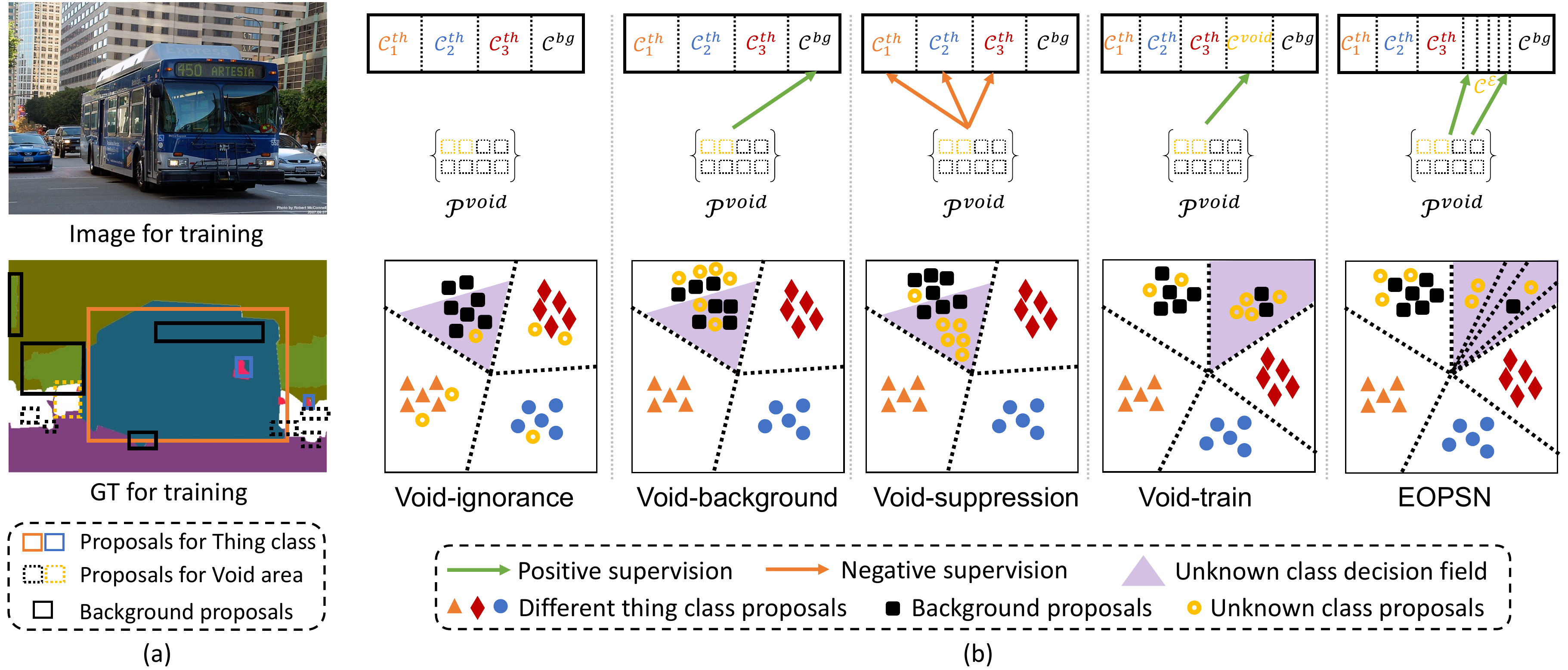}
\caption{Illustration of existing OPS methods. (a) Proposal examples used in existing OPS methods. (b) Display of the usage of``void'' class proposals in different OPS methods (top row) and how \unknown class are predicted at testing phase (bottom row). $\mathcal{C}^{th}_i$, $\mathcal{C}^{bg}$, $\mathcal{C}^{void}$ and $\mathcal{C}^{\mathcal{E}}$ represent classifiers for thing classes, background, ``void'' class and exemplar-based classes, respectively. $\mathcal{P}^{void}$ means ``void'' class proposals. Positive supervision encourages model to produce a higher prediction score while negative supervision prefers a lower score.
}
\label{fig:baseline_decision_boundary}
\end{figure*}
\section{Open-Set Panoptic Segmentation}
\label{sec:bg}
According to the problem definition given in~\cite{hwang2021exemplar}, \textit{open-set panoptic segmentation} (OPS) has
a similar definition to the standard closed-set panoptic segmentation except for the label space and targets of the task. 
Apart from the \known label space (i.e., countable objects in \emph{thing} class $\mathcal{C}^{\text{Th}}$ and amorphous and uncountable regions in \emph{stuff} class $\mathcal{C}^{\text{St}}$) which has annotations at training phase and requires to be effectively segmented during testing, OPS also requires the model to be able to detect and generate instance masks for \emph{unknown thing} class $\mathcal{C}^{\text{Th}}_{u}$ in test set\footnote{Segmentation of \unknown \stuff class is not required in the current OPS definition~\cite{hwang2021exemplar}.}. The \unknown \emph{thing} $\mathcal{C}^{\text{Th}}_{u}$ are not annotated or even not appeared in the training images. For pixel areas in the ground truth of training images that are not manually annotated, a semantic label named \emph{void} will be assigned to them.

\begin{table*}[t]\small 
    \centering
    \begin{tabular}{l|c|c|c|c|c}
        \toprule
        Model & PQ & SQ & RQ & Recall & Precision \\
        \hline
        Void-ignorance & 3.7 & 71.8 & 5.2 & 11.0 & 3.4 \\
        Void-background & 4.3 & 70.1 & 6.2	& 11.2 & 4.2 \\
        Void-suppression & 7.2 & 75.3 & 9.5 & 27.6 & 5.8 \\
        Void-train & 7.5 & 72.9 & 10.3 & 21.8 & 6.7 \\
        EOPSN & 11.3 & 73.8 & 15.3 & 11.8 & 21.9 \\
        \bottomrule
    \end{tabular}
    \vspace{0.2cm}
    \caption{Comparisons of OPS results of \unknown class for existing OPS methods.\protect\footnotemark Since the recognition quality (RQ) varies a lot among these algorithms, recall and precision statistics of \unknown class are also reported for detailed inspection. All empirical numbers are obtained on COCO \emph{val} set with 20\% of \thing classes are set as \unknown class during model training.
    }
    \label{tab:motivation}
\end{table*}
\footnotetext{Results of \known class are comparable among these methods and can be found in Table~\ref{tab:main_res}}

Existing OPS methods~\cite{hwang2021exemplar} are built upon the classic Panoptic FPN network~\cite{kirillov2019panoptic} and this is due to that region proposal network~\cite{ren2015faster} (RPN), an important part of the network, can generate class-agnostic proposals and enable the possible of finding various classes of objects in any image~\cite{gu2022openvocabulary} and makes the OPS problem tractable.

 Figure~\ref{fig:baseline_decision_boundary} (a) presents some proposal examples generated from RPN module where solid boxes in orange and blue denote the proposals are labeled as a specific \known \thing class. Dashed boxes in black and orange denote the ``void'' class proposals\footnote{Proposals who have a half of the region is inside the ``void'' area.} and other black solid boxes are background proposals. Since the proposal labeling of \known classes is based on the \known class GT, the quality of selected proposals are guaranteed. However, the quality of proposals $\mathcal{P}^{void}$ varies greatly as the connected ``void'' area is not manually annotated and may contain multiple objects or just ambiguous pixels, therefore some of them should be labeled as background in the closed-set PS setting. Examples in Figure~\ref{fig:baseline_decision_boundary}(a) show that few yellow dashed boxes are well aligned with an \unknown instance in the ``void'' area, while a large number of black dashed boxes are not well aligned with a specific \unknown instance which should have been labeled as background proposals in the closed-set setting but it is impossible for the open-set case. 

The existing OPS methods differ in how to use ``void'' class proposals and top row of Figure~\ref{fig:baseline_decision_boundary}(b) presents their usage ways: Void-ignorance baseline does not include ``void'' class proposals $\mathcal{P}_{void}$ into network training; Void-background takes $\mathcal{P}_{void}$ as background; Void-suppression alternatively utilizes $\mathcal{P}_{void}$ to do a suppression on \emph{known} class classifiers\footnote{We empirically find that suppress background as well will deteriorate the recognition of \known classes.};
Void-train treats all $\mathcal{P}_{void}$ as the same and adds an \emph{void} class classifier during training; EOPSN method can be seen as an enhanced version of Void-train and builds multiple representative exemplars from $\mathcal{P}_{void}$ through $k$-means clustering.
During testing, proposals will be predicted as \unknown class only when they are rejected by \known classes with a pre-defined confidence threshold. Void-train and EOPSN further require the proposals to be predicted as ``void'' class or exemplar-based classes. Bottom row of Figure~\ref{fig:baseline_decision_boundary}(b) gives a visualization of \unknown class decision field for these methods and their \unknown class recognition quality are presented in Table~\ref{tab:motivation}. We can find that neither Void-ignorance nor Void-background can produce a reasonable \unknown class recognition result. Although both of Void-suppression and Void-train share a similar performance, i.e., relatively high recall and low precision, they may have different reasons. Void-suppression is due to the lack of ability for distinguish \unknown class from background, while Void-train is because the supervision of $\mathcal{P}_{void}$ will make it overfit to the training set. EOPSN greatly improves the precision but heavily affects the recognition recall which means the exemplars obtained from proposals $\mathcal{P}_{void}$ are not representative enough.

\begin{figure*}[t]\footnotesize
\centering
\setlength{\abovecaptionskip}{0.cm}
\includegraphics[scale=0.34]{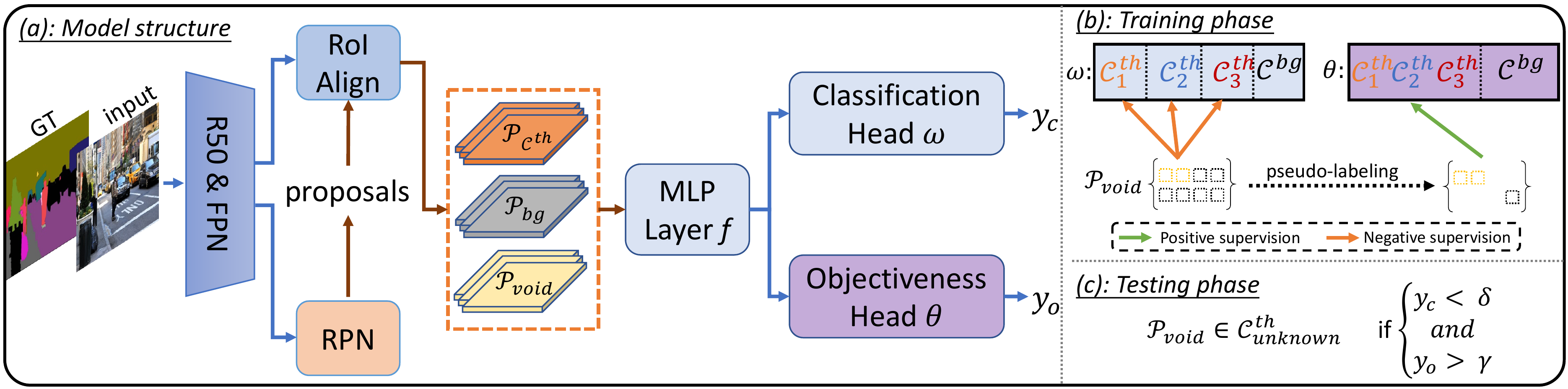}
\caption{Demonstration of our method. We introduce an objectiveness head besides the default classification head for the prediction of objectiveness score of proposals. 
$\mathcal{P}_{\mathcal{C}^{th}}, \mathcal{P}_{bg}$ and $\mathcal{P}_{void}$ represent proposals in \known \thing classes, background and \emph{void} class, respectively. 
}
\label{fig:main_structure}
\end{figure*}

\section{Our Approach}\label{sec:our_approach}
In this section, we first present the necessary of constructing a two-stage decision structure for the OPS task. Then, we further propose a pseudo-labeling method to enhance the generalization ability of \unknown class recognition.

\subsection{Dual Decision Structure for the OPS Task}\label{sec:objhead}
Based on the analysis in Sec.~\ref{sec:bg}, we believe that \unknown class cannot be well modeled at training phase without being aware of what kinds of \unknown class will appear during testing. Therefore, both of Void-train and EOPSN may not be a promising direction for solving the OPS problem and the empirical results on \unseen class~\footnote{\unseen means the corresponding thing never appears in the training images. Sec.~\ref{sec:exp_detail} gives a detail definition.} in Table~\ref{tab:main_3split} confirm our conclusion. However, other OPS methods can only rely on the \known class classifier when making decisions on \unknown classes and the empirical results show that such 
a decision procedure can not enable them to achieve a satisfactory recognition performance for unknown classes. Thus we build up a dual decision process for the effective recognition of \unknown classes.

Following existing OPS methods, our structure is also adapted from the Panoptic FPN framework~\cite{kirillov2019panoptic} and the core structure is presented in Figure~\ref{fig:main_structure}(a). Specifically, for a given image, we first use the ResNet50 network and feature pyramid network to extract multi-scale feature representations. Then a region proposal network is used to generate class-agnostic proposals and their features can be obtained through the RoI align module.
Given the ground truth segmentation annotations at training stage, these proposals can be assigned labels according to their positional relationship with the annotations. For example, the proposals will be labeled as one \known \thing class $\mathcal{C}_i^{th}\in\mathcal{C}^{th}$ when it has a large overlap to any \known \thing class instance. Similarly, the ``void'' areas are also utilized for defining ``void'' class proposals $\mathcal{P}_{\emph{void}}$. Other proposals are labeled as background class samples $\mathcal{P}_{bg}$. 

In order to identify \known classes, the classification head is supervised by the proposals
\begin{equation}\label{eq:cls_loss}
    \min - \frac{1}{N_{\mathcal{P}_{\overline{\emph{void}}}}} \sum_{i\in\{\mathcal{C}^{th}, bg\}} \sum_{k=1}^{N_{\mathcal{P}_i}}\log \frac{\exp\bigl (w_i^T f(\mathcal{P}_i^k) \bigr)}{\sum_{j\in \{\mathcal{C}^{\text{Th}}, bg\}}\exp\bigl(w_j^T f(\mathcal{P}_i^k)\bigr)}
\end{equation}
where $w$ is the weight of classification head. $N_{\mathcal{P}_{\overline{\emph{void}}}}$ is the number of proposals except for those belonging to ``void'' class. $N_{\mathcal{P}_i}$ is the number of proposals in any specific \thing class. In order to separate \known and \unknown class effectively, we follow the Void-suppression baseline to do a suppression on \known class classifiers with ``void'' class proposals
\begin{equation}\label{eq:void_supp}
    \min - \frac{1}{N_{\mathcal{P}_{void}}} \sum_{i=1}^{N_{\mathcal{P}_{void}}} \sum_{k\in \mathcal{C}^{\text{th}}}\log \Bigl(1 - \frac{\exp\bigl (w_k^T f(\mathcal{P}_{void}^i) \bigr)}{\sum_{\{\mathcal{C}^{\text{th}}, bg\}}\exp\bigl(w_j^T f(\mathcal{P}_{void}^i)\bigr)} \Bigr).
\end{equation}where $N_{\mathcal{P}_{\emph{void}}}$ means the number of ``void'' class proposals.

Considering the modeling of Eqs.~\ref{eq:void_supp} and~\ref{eq:cls_loss} can only improve the discriminative ability of \known classes, \unknown classes may still mix with background ones. In order to mitigate this drawback, 
we introduce a class-agnostic object prediction head (a.k.a. objectiveness head) parallel to the \known class classification head and optimize it as follows
\begin{equation}\label{eq:obj_loss}
    \min \frac{-1}{N_{\mathcal{P}_{\overline{\emph{void}}}}} \biggl( \sum_{i\in\mathcal{C}^{th}} \sum_{k=1}^{N_{\mathcal{P}_i}}\log \frac{\exp\bigl (\theta^T f(\mathcal{P}_i^k) \bigr)}{1 + \exp\bigl (\theta^T f(\mathcal{P}_i^k) \bigr)} + \sum_{l=1}^{N_{\mathcal{P}_{bg}}} \log \frac{1}{1 + \exp{\bigl(\theta^T f(\mathcal{P}_{bg}^l)\bigr)}} \biggr )
\end{equation}
where $\theta$ is the weight of objectiveness head and $N_{\mathcal{P}_{bg}}$ is the number of background proposals. 

At the testing stage, the recognition of \unknown class will be made in a dual decision process based on the predictions on both \known class classification head and class-agnostic object prediction head, i.e., only proposals who are rejected by the \known class classification head and accepted by the objectiveness head simultaneously will be predicted as \unknown class. Empirical results in Table~\ref{tab:main_res} shows that such kind of dual decision process significantly boosts the \unknown class recognition performance on all kinds of OPS settings.

\smallskip
\noindent{\textbf{Rationale of design:}} The key feature of the above design is that we will treat all known class proposals as training samples for \textit{a single class-agnostic} ``object'' class. In contrast, the methods described in Figure \ref{fig:baseline_decision_boundary} will treat each class separately. The class-agnostic classification head will encourage the network identify patterns that are shared across class rather than focusing on (known-)class specific patterns. The former can generalize well to unseen thing while the latter may overfit to things only seen at the training stage. 

\subsection{Improve Object Recognition Generalization with Pseudo-labeling}
\label{sec:pl_void}
Currently, the newly added class-agnostic object prediction head is only optimized on proposals belonging to \known \thing class or background ones and the ``void'' class proposals~\footnote{We take any connected ``void'' area in ground truth of training images as ``void'' class proposals.} are not fully utilized. Since the proposals of ``void'' class may contain many novel objects which does not belong to the annotated \known \thing classes, we assume the properly exploiting of ``void'' class proposals can be helpful for the recognition generalization of objectiveness head. One straightforward way is to directly take all the ``void'' class proposals as potential \unknown ones to supervise the objectiveness head. However, results in 
Figure~\ref{fig:abl_confid} shows that this strategy will heavily deteriorate the recognition quality. It may because the proposals of \emph{void} class are not precise and contain much noise which is not suitable for the immediate exploiting. Therefore, we propose to use the pseudo-labeling technique to filter out invalid ``void'' class proposals. 

Since the newly added objectiveness head is designed in a class-agnostic fashion, the quality of ``void'' class proposals can be predicted by the up-to-date objectiveness head and we can select those high confident ones to further supervise the objectiveness head
\begin{equation}\label{eq:obj_loss_online_pl}
    \min \ \frac{-1}{N_{\mathcal{P}_{void}}} \sum_{i=1}^{N_{\mathcal{P}_{void}}} \mathbbm{1}{\biggl(\frac{\exp\bigl(\theta^T f(\mathcal{P}_{void}^i)\bigr)}{1 + \exp\bigl(\theta^T f(\mathcal{P}_{void}^i)\bigr)} \geq \delta\biggr)} \log \frac{\exp\bigl(\theta^T f(\mathcal{P}_{void}^i)\bigr)}{1 + \exp\bigl(\theta^T f(\mathcal{P}_{void}^i)\bigr)} \\
\end{equation}
where $\delta$ is the confidence threshold.

\begin{table*}[t] 
\centering
\scalebox{0.75}{%
\setlength{\tabcolsep}{4pt}
\begin{tabular}{cl|ccc|ccc|ccc|ccc|cc}
\toprule
\multirow{2}{*}{$K$} & \multirow{2}{*}{Model} &  \multicolumn{9}{c|}{Known} & \multicolumn{5}{c}{Unknown} \\ \cline{3-16}
&  & \addstackgap[3.5pt]{PQ} & SQ & RQ & $\text{PQ}^{\text{Th}}$ & $\text{SQ}^{\text{Th}}$& $\text{RQ}^{\text{Th}}$ & $\text{PQ}^{\text{St}}$ & $\text{SQ}^{\text{St}}$ & $\text{RQ}^{\text{St}}$ & PQ & SQ & RQ & R & P \\ \hline
 & Supervised & 39.4 & 77.7 & 48.4 & 45.8 & 80.7 & 55.4 & 29.7 & 73.1 & 38.0 &- &- &- &- &- \\ \hline
\multirow{4}{*}{5} & Void-supp. & 38.0 & 77.0 & 46.7 & 44.8 & 80.6 & 54.1 & 28.3 & 71.7 & 36.1 & 6.7 & 76.2 & 8.8 & 39.9 & 4.9 \\
& Void-train & 37.3 & 76.7 & 45.9 & 43.6 & 80.4 & 52.8 & 28.2 & 71.5 & 36.0 & 8.6 & 72.7 & 11.8 & 29.8 & 7.3 \\
& EOPSN & 38.0 & 76.9 & 46.8 & 44.8 & 80.5 & 54.2 & 28.3 & 71.9 & 36.2 & 23.1 & 74.7 & 30.9 & 25.9 & 38.3\\ 
& \textbf{Ours} & 38.1 & 77.7 & 46.6 & 45.1 & 80.9 & 54.3 & 28.1 & 73.1 & 35.7 & \textbf{30.2} & \textbf{80.0} & \textbf{37.8} & 32.8 & 44.5 \\
\hline

\multirow{4}{*}{10} & Void-supp. & 37.6 & 76.8 & 46.3 & 44.3 & 80.5 & 53.5 & 28.5 & 71.7 & 36.4 & 6.5 & 76.0 & 8.6 & 32.7 & 5.0 \\
& Void-train & 37.1 & 77.1 & 45.8 & 43.7 & 80.1 & 53.1 & 28.1 & 73.0 & 35.9 & 8.1 & 72.6 & 11.2 & 25.7 & 7.2 \\
& EOPSN & 37.7 & 76.8 & 46.3 & 44.5 & 80.6 & 53.8 & 28.4 & 71.8 & 36.2 & 17.9 & 76.8 & 23.3 & 19.0 & 30.2	\\
& \textbf{Ours} & 37.7 & 77.1 & 46.3 & 45.0 & 80.7 & 54.3 & 27.8 & 72.2 & 35.4 & \textbf{24.5} & \textbf{79.9} & \textbf{30.7} & 24.7 & 40.6 \\
\hline

\multirow{4}{*}{20} & Void-supp. & 37.5 & 75.9 & 46.1 & 45.1 & 80.6 & 54.5 & 28.2 & 70.2 & 36.1 & 7.2 & 75.3 & 9.5 & 27.6 & 5.8 \\
& Void-train & 36.8 & 76.3 & 45.4 & 44.1 & 80.1 & 53.5 & 27.9 & 71.6 & 35.6 & 7.5 & 72.9 & 10.3 & 21.8 & 6.7 \\
& EOPSN & 37.4 & 76.2 & 46.2 & 45.0 & 80.3 & 54.5 & 28.2 & 71.2 & 36.2 & 11.3 & 73.8 & 15.3 & 11.8 & 21.9  \\
& \textbf{Ours} & 37.1 & 75.8 & 45.7 & 45.0 & 80.6 & 54.3 & 27.6 & 70.1 & 35.3 & \textbf{21.4} & \textbf{79.1} & \textbf{27.1} & 21.9 & 35.4\\
\bottomrule
\end{tabular}
}
\vspace{0.2cm}
\caption{
Comparisons of open-set panoptic segmentation performance against the state-of-the-art methods on MS-COCO \val set with three \known-\unknown splits $K$(\%) which denotes the ratio of \unknown classes to all classes. Recall (R) and precision (P) of \unknown objects are also presented for reference. The best results on \unknown classes are bold highlighted.
}
\label{tab:main_res}
\end{table*}
\begin{table*}[t]
\centering
\scalebox{0.75}{%
\setlength{\tabcolsep}{4pt}
\begin{tabular}{@{}l|ccc|ccc|ccc|ccc|ccc@{}}
\toprule
\multirow{2}{*}{Model} & \multicolumn{9}{c|}{Known} & \multicolumn{3}{c|}{Unknown} & \multicolumn{3}{c}{Unseen}\\ \cline{2-16}
  & \addstackgap[3.5pt]{PQ} & SQ &RQ & $\text{PQ}^{\text{Th}}$ & $\text{SQ}^{\text{Th}}$& $\text{RQ}^{\text{Th}}$ & $\text{PQ}^{\text{St}}$ & $\text{SQ}^{\text{St}}$ & $\text{RQ}^{\text{St}}$ & PQ & SQ &RQ & PQ & SQ &RQ\\ 
\hline
Void-supp. & 35.8 & 76.7 & 44.5 & 43.0 & 81.2 & 52.5 & 27.7 & 71.6 & 35.4 & 7.6 & 75.5 & 10.1 & 4.5 & 75.9 & 6.0\\
Void-train & 35.4 & 77.2 & 43.9 & 42.2 & 81.0 & 51.6 & 27.7 & 72.8 & 35.3 & 8.8 & 73.8 & 15.7 & 4.4 & 74.8 & 5.9\\
EOPSN & 35.7 & 76.6 & 44.7 & 43.2 & 81.1 & 52.7 & 27.8 & 71.4 & 35.6 & 23.0 & 74.6 & 30.8 & 0.4 & 80.3 & 0.5 \\
\textbf{Ours} & 35.8 & 76.6 & 44.5 & 43.0 & 81.1 & 52.5 & 27.6 & 71.4 & 35.3 & \textbf{30.2} & \textbf{80.2} & \textbf{37.7} & \textbf{9.3} & \textbf{82.5} & \textbf{11.2} \\
\bottomrule
\end{tabular}
} 
\vspace{0.2cm}
\caption{Comparisons of OPS performance on MS-COCO \val set under the newly proposed \emph{zero-shot} setting. The best results on \unknown and \emph{unseen} classes are bold highlighted.}
\label{tab:main_3split}
\end{table*}

\section{Experimental Results}\label{sec:exp_exp}
In this section, we conduct experiments to evaluate the proposed approach and existing OPS methods on open-set panoptic segmentation task.

\subsection{Experimental Details}\label{sec:exp_detail}
To make a fair comparison, we directly build up our experiments based on the released codebase\footnote{\url{https://github.com/jd730/EOPSN.git}}.
Some experimental details are as follows

\smallskip
\noindent \textbf{Datasets}: Following the protocol of \cite{hwang2021exemplar}, all experiments are conducted on MS-COCO 2017 dataset whose default annotations are constructed by 80 \thing classes and 53 \stuff classes. \cite{hwang2021exemplar} manually removes a subset of \known \thing classes (i.e., $K$\% of 80 classes) in the training dataset and takes them as \unknown classes for evaluating on open-set task (\stuff classes are all kept). Three \known-\unknown splits of $K$ are considered:\textbf{5\%}, \textbf{10\%} and \textbf{20\%}.

In order to evaluate the object recognition generalization ability of OPS methods, we further construct a more realistic OPS setting named \textbf{\emph{zero-shot}} which is built up from the 5\% split setting mentioned above and further removes training images that contains instances belonging to the 20\% tail \thing class of MS-COCO. These classes are \{hair drier, toaster, parking meter, bear, scissors, microwave, fire hydrant, toothbrush, stop sign, mouse, refrigerator, snowboard, frisbee, keyboard, hot dog, baseball bat\}. To distinguish from \unknown classes, we call these classes \textbf{\unseen classes}.

\smallskip
\noindent \textbf{Methods}: Two strong baselines and the state-of-the-art OPS method are included for comparison, i.e., Void-train, Void-suppression and EOPSN. Meanwhile, Panoptic FPN trained on full 80 \thing classes are also reported for a reference baseline (denoted as supervised).

\smallskip
\noindent \textbf{Evaluation Metric}: The standard panoptic segmentation metrics (i.e., PQ, SQ, RQ) are reported for \known, \unknown and \emph{unseen} classes (see detail formulations in the appendix).

\begin{figure*}[t]\footnotesize
\centering
\setlength{\abovecaptionskip}{0.cm}
\includegraphics[scale=0.45]{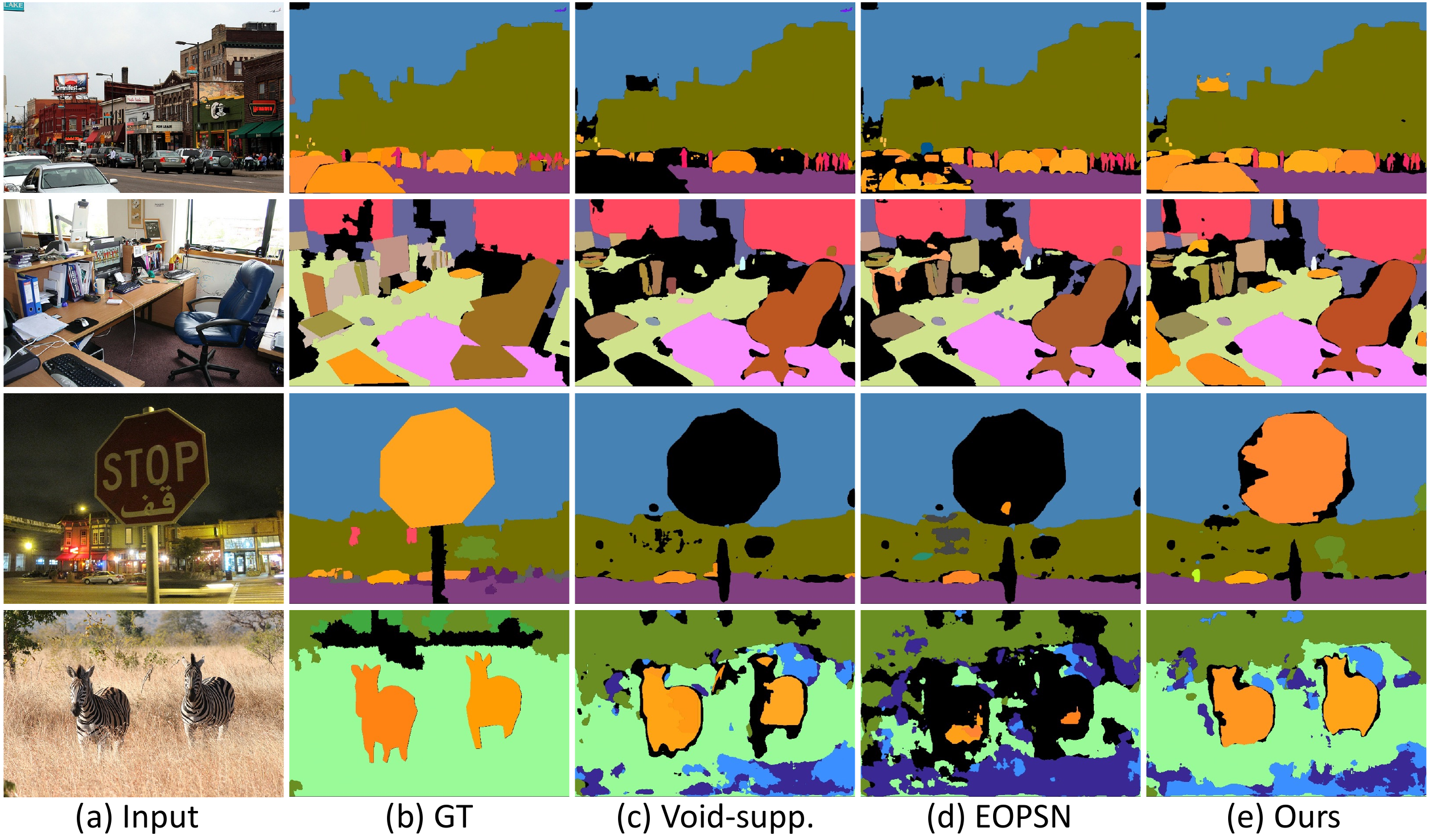}
\caption{Visual results on COCO \val set with $K$=20\%. Compared to Void-suppression and EOPSN, our algorithm can detect more novel objects and generates better instance masks. The most prominent \unknown class objects in row 1-4 are car, keyboard, stop sign and zebra, respectively. (a) Input image (b) Ground-truth (c), (d) and (e) are panoptic segmentation results of Void-suppression, EOPSN and our method, respectively. Predicted instances in the \unknown class are denoted by orange color and the black areas represent the areas that are fail annotated (i.e., (b)) or segmented (i.e., (c)-(e)).
}
\label{fig:exp_k20}
\end{figure*}

\subsection{Results on \known-\unknown Setting}\label{sec:exp_kn_unk}

Table~\ref{tab:main_res} shows the quantitative results of comparing methods. It is clear that our proposed method significantly improves the panoptic quality of \unknown class objects than the Void-suppression baseline across all kinds of splits. Meanwhile, compared with the SOTA method EOPSN, our approach excels on both of the recall and precision of \unknown objects recognition and therefore achieves much better PQ values. 
Figure~\ref{fig:exp_k20} illustrates the qualitative results. We find that our approach can successively detect more \unknown class objects and generate more precise instance masks than both of Void-suppression baseline and EOPSN method.

\begin{figure}
    \begin{minipage}{.55\linewidth}
    \small
    \centering
    \setlength\tabcolsep{3pt}
    \begin{tabular}{lcc|ccc|ccc}
        \toprule
         & \multirow{2}{*}{Obj.} & \multirow{2}{*}{PL} & \multicolumn{3}{c|}{Unknown} & \multicolumn{3}{c}{Unseen} \\ \cline{4-9}
        & & & \addstackgap[0.5pt]{PQ} & SQ & RQ & PQ & SQ & RQ \\ 
        \hline
        \parbox[t]{2mm}{\multirow{3}{*}{\rotatebox[origin=c]{90}{$K$=20\%}}} & \xmark & \xmark & 7.2 & 75.3 & 9.5 & - & - & - \\
        & \cmark & \xmark & 19.5 & \textbf{79.5} & 24.5 & - & - & - \\
        & \cmark & \cmark & \textbf{21.4} & 79.1 & \textbf{27.1} & - & - & - \\
        \hline

        \parbox[t]{2mm}{\multirow{3}{*}{\rotatebox[origin=c]{90}{zero-shot}}} & \xmark & \xmark & 7.6 & 75.5 & 10.1 & 4.5 & 75.9 & 6.0 \\
        & \cmark & \xmark & 29.6 & \textbf{80.5} & 36.8 & 6.9 & 81.3 & 8.5\\
        & \cmark & \cmark & \textbf{30.2} & 80.2 & \textbf{37.7} & \textbf{9.3} & \textbf{82.5} & \textbf{11.2} \\
        
        \bottomrule
    \end{tabular}
    \vspace{0.2cm}
    \captionof{table}{Ablation study to the effectiveness of each component in our method.}
    \label{tab:abl_void_obj_plus}
    \end{minipage}\hspace{.25cm}%
    \begin{minipage}{.4\linewidth}
    \centering
    \setlength{\abovecaptionskip}{0.cm}
    \includegraphics[width=0.8\linewidth]{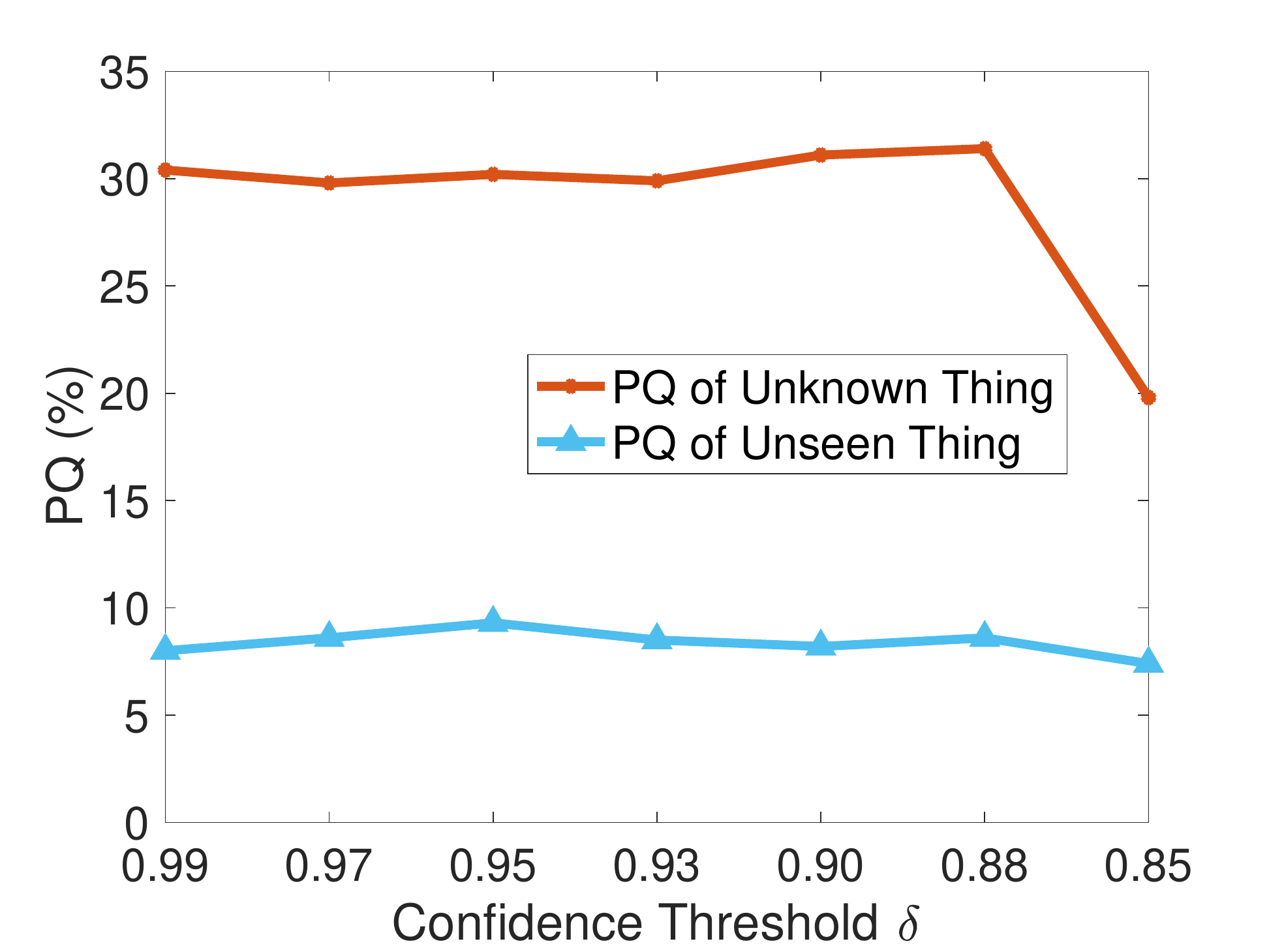}
    \caption{Ablation study of confidence threshold of pseudo labeling on \emph{zero-shot} setting.}
    \label{fig:abl_confid}
    \end{minipage}
\end{figure}

\subsection{Results on \emph{zero-shot} Setting}\label{sec:exp_zeroshot}

Our approach has been verified to be effective on \known-\unknown setting in Sec.~\ref{sec:exp_kn_unk}, we also want to know its novel object recognition ability in a \emph{zero-shot} setting. Table~\ref{tab:main_3split} presents that the proposed methods are superior than the comparing ones on both \unknown class and \unseen class objects. It is interesting that EOPSN performs well on \unknown class but almost fails on \emph{unseen} class. This may be due to the fact that the exemplars obtained in EOPSN are completely derived from the training set and cannot be generalized to unseen class objects. 
Qualitative results for the \emph{zero-shot} setting has been presented in the appendix due to the space limit and our approach can always detect salient objects in the image and produce overall best instance masks.

\subsection{Ablation Study}\label{sec:exp_abl_study}
We are interested in ablating our approach from the following perspective views:

\smallskip
\noindent \textbf{Effective of each component in our method:} 
Our approach is mainly composed of two components (i.e., objectiveness head and pseudo labeling) and Table~\ref{tab:abl_void_obj_plus} shows the performance contribution of each component on two kinds of settings. It is obvious that simply adding the objectiveness head significantly improves the unknown segmentation performance and incorporating the pseudo labeling trick further boost the overall performance.

\smallskip
\noindent \textbf{Sensitivity analysis:} Our method only has one hyper-parameter, i.e., the confidence threshold $\delta$ in pseudo labeling mechanism. As shown in Figure~\ref{fig:abl_confid}, the performance of our approach is stable when the confidence value falls into $\delta \in [0.88, 0.99]$.

\section{Conclusion}
Open-set panoptic segmentation (OPS) is a newly proposed research task which aims to perform segmentation for both \known classes and \unknown classes. In order to solve the challenges of OPS, we propose a dual decision mechanism for \unknown class recognition. We implement this mechanism through coupling a \known class classification head and a class-agnostic object prediction head and make them corporate together for final \unknown class prediction. To further improve the recognition generalization ability of the objectiveness head, we use the pseudo-labeling technique to boost the performance of our approach. Extensive experimental results verify the effectiveness of the proposed approach on various kinds of OPS tasks. 

\clearpage

\bibliography{egbib}
\end{document}


\maketitle

In this appendix, we first present detail formulations of the panoptic segmentation evaluation metrics. We then list details about the dataset splits used in our paper. Finally, we demonstrate the qualitative results for the \emph{zero-shot} setting. 

\section{Formulation Details of Evaluation Metric}
Three kinds of panoptic segmentation metrics are normally considered in the literature, i.e., panoptic quality (PQ), segmentation quality (SQ) and the recognition quality (RQ)
\begin{equation*}
    \text{Panoptic Quality (PQ)} = \underbrace{\frac{\sum_{(p, g) \in \TP} \text{IoU}(p, g)}{\vphantom{\frac{1}{2}}|\TP|}}_{\text{segmentation quality (SQ)}} \cdot \underbrace{\frac{|\TP|}{|\TP| + \frac{1}{2} |\FP| + \frac{1}{2} |\FN|}}_{\text{recognition quality (RQ)}}\\
\end{equation*}
where $\text{IoU}(p, g)$ means intersection over union of predicted $p$ and ground truth $g$ segments. $\TP$/$\FP$/$\FN$ respectively denote the set of true positives, false positives and false negatives. For the open-set panoptic segmentation task, we use these three metrics to evaluate the performance of \known, \unknown and \unseen classes.

\section{Details of Dataset Splits}
In our paper, we have conducted experiments on two kinds of open-set panoptic segmentation settings, i.e., \known-\unknown setting and \emph{zero-shot} setting. For the \known-\unknown setting, three kinds of splits are evaluated following the previous work EOPSN~\cite{hwang2021exemplar}, i.e., K\% of classes are removed from the 80 \thing classes of MS-COCO 2017 as \unknown classes and $K=\{5, 10, 20\}$. We list the \unknown classes as follows (the classes are removed cumulatively for these three settings respectively)
\begin{itemize}
    \item car, cow, pizza,  toilet
    \item boat, tie, zebra,  stop sign
    \item dining table, banana, bicycle, cake, sink, cat, keyboard, bear
\end{itemize}

For the more realistic \emph{zero-shot} setting, we build it upon the 5\% split \known-\unknown setting mentioned above and further removes training images that contains instances belonging to the 20\% tail \thing class of MS-COCO. The removed tail classes has already been presented in the main paper.

\section{Qualitative Results of \emph{zero-shot} Setting}

\begin{figure*}[h]\footnotesize
\centering
\setlength{\abovecaptionskip}{0.cm}
\includegraphics[scale=0.45]{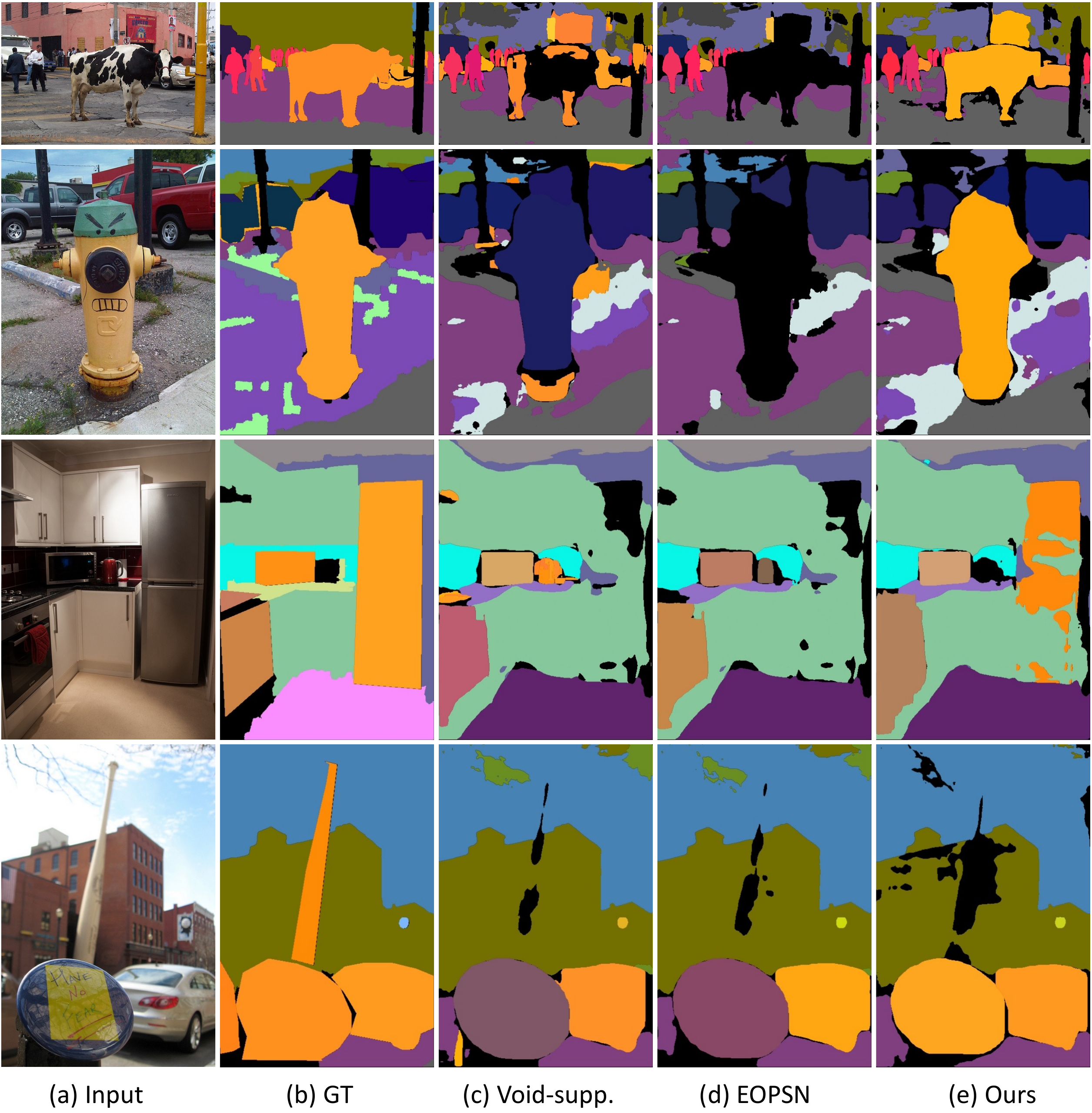}
\caption{Visual results on COCO \val set with \unknown and \unseen class. The first row shows that our method can generate better instance mask for \unknown class, e.g., cow. Row 2-4 present that the proposed approach can successfully detect some \unseen classes (tail classes on COCO), e.g., fire hydrant, refrigerator and parking meter. 
}
\label{fig:exp_3split}
\end{figure*}

In the main paper, we have presented the superior quantitative results of our approach on the \emph{zero-shot} setting. In the appendix, we further visualize the qualitative results of our approach on the \emph{zero-shot} setting. As Figure~\ref{fig:exp_3split} shows, our approach can not only detect more unknown class objects but also generate more precise instance masks than both of Void-suppression baseline and EOPSN method.

\clearpage

\bibliography{egbib}